\documentclass[10pt,conference]{IEEEtran}
\IEEEoverridecommandlockouts

\usepackage[numbers]{natbib}
\usepackage{amsmath,amssymb,amsfonts}
\usepackage{booktabs}
\usepackage{paralist}
\usepackage{tcolorbox}
\usepackage[linesnumbered, ruled, vlined]{algorithm2e}

\SetAlFnt{\small}
\SetAlCapFnt{\small}
\SetAlCapNameFnt{\small}
\SetKwRepeat{Do}{do}{while}

\definecolor{lightgray}{rgb}{0.83, 0.83, 0.83}

\usepackage{pgfplotstable}
\newcolumntype{R}{>{\raggedleft\arraybackslash}p{3mm}}

\usepackage{hyperref}

\def\BibTeX{{\rm B\kern-.05em{\sc i\kern-.025em b}\kern-.08em
    T\kern-.1667em\lower.7ex\hbox{E}\kern-.125emX}}
\begin{document}

\title{Many-Objective Reinforcement Learning for Online Testing of DNN-Enabled Systems
\thanks{
This work has been carried out as part of the COSMOS Project, which has received funding from the European Union’s Horizon 2020 Research and Innovation Programme under grant agreement No. 957254. This work was also supported by NSERC of Canada under the Discovery and CRC programs. 
}
}

\author{
	\IEEEauthorblockN{Fitash Ul Haq}
	\IEEEauthorblockA{\textit{University of Luxembourg}\\
		Luxembourg\\
		fitash.ulhaq@uni.lu}
	\and
	\IEEEauthorblockN{Donghwan Shin\IEEEauthorrefmark{1}\thanks{\IEEEauthorrefmark{1}Part of this work was done while the author was affiliated with the University of Luxembourg, Luxembourg.}}
	\IEEEauthorblockA{\textit{University of Sheffield}\\
		Sheffield, United Kingdom\\
		d.shin@sheffield.ac.uk}
	\and
	\IEEEauthorblockN{Lionel C. Briand}
	\IEEEauthorblockA{\textit{University of Luxembourg}\\
		Luxembourg\\
		\textit{University of Ottawa}\\
		Ottawa, Canada\\
		lionel.briand@uni.lu}
}

\maketitle

\begin{abstract}
Deep Neural Networks (DNNs) have been widely used to perform real-world tasks in cyber-physical systems such as Autonomous Driving Systems (ADS). 
Ensuring the correct behavior of such DNN-Enabled Systems (DES) is a crucial topic.
Online testing is one of the promising modes for testing such systems with their application environments (simulated or real) in a closed loop, taking into account the continuous interaction between the systems and their environments.
However, the environmental variables (e.g., lighting conditions) that might change during the systems' operation in the real world, causing the DES to violate requirements (safety, functional), are often kept constant during the execution of an online test scenario due to the two major challenges:
(1) the space of all possible scenarios to explore would become even larger if they changed and
(2) there are typically many requirements to test simultaneously.

In this paper, we present MORLOT (Many-Objective Reinforcement Learning for Online Testing), a novel online testing approach to address these challenges by combining Reinforcement Learning (RL) and many-objective search.
MORLOT leverages RL to incrementally generate sequences of environmental changes while relying on many-objective search to determine the changes so that they are more likely to achieve any of the uncovered objectives. 
We empirically evaluate MORLOT using CARLA, a high-fidelity simulator widely used for autonomous driving research, integrated with Transfuser, a DNN-enabled ADS for end-to-end driving. 
The evaluation results show that MORLOT is significantly more effective and efficient than alternatives with a large effect size. 
In other words, MORLOT is a good option to test DES with dynamically changing environments while accounting for multiple safety requirements. 
\end{abstract}


\section{Introduction}\label{sec:intro}
Deep Neural Networks (DNNs) have been widely used to perform various tasks, such as object classification~\cite{krizhevsky2012imagenet}, speech recognition~\cite{dahl2011context} and object detection~\cite{szegedy2013deep}. 
With the recent advances in Deep Learning (DL), DNNs are increasingly applied to safety-critical software systems, such as Automated Driving Systems (ADS) that drive a vehicle with the aim of preventing safety and functional violations (e.g., colliding with other vehicles).
Therefore, ensuring the correct behavior of such DNN-Enabled Systems (DES) has emerged as a fundamental problem of software testing.

Online testing~\cite{9000651} is one of the promising modes of DES testing that accounts for the closed-loop interaction between the DES under test and its environment.
In online testing, the DES under test is embedded into the application environment (e.g., a simulated driving environment for ADS) and is monitored to detect the violations of safety and functional requirements  during the closed-loop interaction.
Such interaction helps account for the accumulation of (possibly small) errors over time, eventually leading to requirements violations that often cannot be detected by offline testing~\cite{fitash-journal}. As a result, online testing has been actively applied in many testing approaches recently~\cite{asfault,ul2022efficient,riccio2020model,koren2018adaptive,lu2022learning}.

However, existing online testing approaches for DES exhibit at least one of two critical limitations.
First, they do not account for the fact that there are often many safety and functional requirements, possibly independent of each other, that must be considered together in practice. 
Though one could simply repeat an existing test approach for individual requirements, it is inefficient due to its inability to dynamically distribute the test budget (e.g., time) over many requirements according to the feasibility of requirements violations. 
For example, if one of the requirements cannot be violated, the pre-assigned budget for this requirement would simply be wasted. 
Furthermore, dividing the limited test budget across many requirements may result in too small a budget for testing individual requirements thoroughly. 
Second, they do not vary dynamic environmental elements, such as neighboring vehicles and weather conditions, during test case (i.e., test scenario) execution.
For example, certain weather conditions (e.g., sunny) remain the same throughout a test scenario, whereas in reality they may change over time, which can trigger requirements violations. 
This is mainly because the number of possible test scenarios increases exponentially when considering individual dynamic elements' changes (i.e., time series).
However, not accounting for such dynamic environments could significantly limit test effectiveness by limiting the scope of the test scenarios being considered.

\textit{\textbf{Contributions}}.
To overcome the above limitations, we present MORLOT (Many-Objective Reinforcement Learning for Online Testing), a novel online testing approach for DES.
MORLOT leverages two distinct approaches: (1) Reinforcement Learning (RL)~\cite{sutton2018reinforcement} to generate the sequences of changes to the dynamic elements of the environment with the aim of causing requirements violations, and (2) many-objective search~\cite{mosa,fitest} to efficiently satisfy as many independent requirements as possible within a limited testing budget.
The combination of both approaches works as follows: 
(1) RL incrementally generates the sequence of changes for the dynamic elements of the environment, and 
(2) those changes are determined by many-objective search such that they are more likely to achieve any of the uncovered objective (i.e., the requirement not yet violated). 
In other words, changes in the dynamic elements tend to be driven by the objectives closest to being satisfied. 
For example, if there are three objectives $o_1$, $o_2$, and $o_3$ where $o_2$ is closest to being satisfied, MORLOT incrementally appends those changes that help in achieving $o_2$ to the sequence.
Furthermore, by keeping a record of uncovered objectives as the state-of-the-art many-objective search for test suite generation does, MORLOT focuses on the uncovered ones over the search process.

Though MORLOT can be applied to any DES interacting with an environment including dynamically changeable elements, it is evaluated on DNN-enabled Autonomous Diving Systems (DADS). 
Specifically, we use Transfuser~\cite{Prakash2021CVPR}, the highest ranked DADS, at the time of our evaluation, among publicly available ones in the CARLA Autonomous Driving Leaderboard~\cite{carla-leaderboard} and CARLA \cite{CARLA-citation}, a high-fidelity simulator that has been widely used for training and validating DADS.
Our evaluation results, involving more than 600 computing hours, show that MORLOT is significantly more effective and efficient at finding safety and functional violations than state-of-the-art, many-objective search-based testing approaches tailored for test suite generation~\cite{mosa,fitest} and random search.

Our contributions can be summarized as follows:
\begin{itemize}[-]
    \item MORLOT, a novel approach that efficiently generates complex test scenarios, including sequential changes to the dynamic elements of the environment of the DES under test, by leveraging both RL and many-objective search;
    \item An empirical evaluation of MORLOT in terms of test effectiveness and efficiency and comparison with alternatives;
    \item A publicly available replication package, including the implementation of MORLOT and instructions to set up our case study.
\end{itemize}

\textit{\textbf{Significance}}.
For DES that continuously interact with their operational environments and have many safety and functional requirements to be tested, performing online testing efficiently to identify as many requirements violations as possible, without arbitrarily limiting the space of test scenarios, is essential. 
However, taking into account the sequential changes of dynamic elements of an environment over time is extremely challenging since it renders the test scenario space exponentially larger as test scenario time increases. 
MORLOT addresses the problem by leveraging and carefully combining RL and many-objective search, thus providing an important and novel contribution towards scalable online testing of real-world DES in practice.

\textit{\textbf{Paper Structure}}.
The rest of the paper is structured as follows. 
Section~\ref{sec:background} provides a brief background on RL and many-objective search. 
Section~\ref{sec:problem} formalizes the problem of DES online testing with a dynamically changing environment. 
Section~\ref{sec:related-work} discusses and contrasts related work. 
Section~\ref{sec:approach} describes our proposed approach, starting from a generic RL-based test generation approach and then MORLOT, a many-objective reinforcement learning approach for online testing. 
Section~\ref{sec:eval} evaluates the test effectiveness and efficiency of MORLOT using an open-source DNN-based ADS with a high-fidelity driving simulator.
Section~\ref{sec:conclusion} concludes the paper.
Section~\ref{sec:data} provides details on the replication package.
\section{Background}\label{sec:background}

\subsection{Reinforcement Learning}\label{sec:background:rl}
Reinforcement Learning (RL) is about learning how to perform a sequence of actions to achieve a goal by iterating trials and errors to learn the best action for a given state~\cite{sutton2018reinforcement}.

In particular, RL involves the interaction between an RL agent (i.e., the learner) and its surrounding environment, which is formalized by a Markov Decision Process (MDP).
At each time step $j$, the RL agent observes the environment's state $s_j$ and takes an action $a_j$ based on its own policy $\pi$ (i.e., the mapping between states and actions).
At the next time step $j+1$, the agent first gets a reward $w_{j+1}$ indicating how well taking $a_j$ in $s_j$ helped in achieving the goal, updates $\pi$ based on $w_{j+1}$, and then continues to interact with its environment. 
From the interactions (trials and errors), the agent is expected to learn the unknown optimal policy $\pi^*$ that can select the best action maximizing the expected sum of future rewards in any state.

An important assumption underlying MDP is that states satisfy the \emph{Markov property}: states captures information about all aspects of the past agent–environment interactions that make a difference for the future~\cite{sutton2018reinforcement}. 
In other words, $w_{j+1}$ and $s_{j+1}$ depend only on $s_j$ and $a_j$, and are independent from the previous states $s_{j-1}, s_{j-2}, \dots, s_1$ and actions $a_{j-1}, a_{j-2}, \dots, a_1$. 
This assumption allows the RL agent to take an action by considering only the current state, as opposed to all past states (and actions). 

In general, there are two types of RL methods: (1) \emph{tabular}-based and (2) \emph{approximation}-based.
Tabular-based methods~\cite{watkins1992q,rummery1994line} use tables or arrays to store the expected sum of future rewards for each state. 
Though they are applicable only if state and action spaces are small enough to be represented in tables or arrays, they can often find exactly the optimal policy~\cite{sutton2018reinforcement}.
Discretization can be used to control the size of state and action spaces, especially when states and actions are continuous.
It is essential to apply the right degree of discretization since coarse-grained discretization may make it impossible for the agent to distinguish between states that require different actions, resulting in significant loss of information.
When the state space is enormous and cannot be easily discretized without significant information loss, approximation-based methods~\cite{8103164} can be used where the expected sum of future rewards for a newly discovered state can be approximated based on known states (often with the help of state abstraction when they are too complex to directly compare). 
While they can address complex problems in very large state spaces, they can only provide approximate solutions.

One of the most commonly used tabular-based reinforcement learning algorithms is Q-learning due to its simplicity and guaranteed convergence to an optimal policy~\cite{watkins1992q}. 
It stores and iteratively updates the expected sum of future rewards for each state-action pair in a table (a.k.a., Q-table) while going through trials and errors. 
A properly updated Q-table can therefore tell what is the best action to choose in a given state.
A more detailed explanation, including how to update a Q-table to ensure the convergence, can be found in \citet{sutton2018reinforcement}.

\subsection{Many-Objective Search}\label{sec:background:many-search}
Many-objective search~\cite{4631121,chand2015evolutionary} refers to solving multiple objectives simultaneously, typically more than four, using meta-heuristic search algorithms, such as MOEA/D~\cite{4358754} and NSGA-III~\cite{nsga-iii}. 
With the help of fitness functions that quantify the goodness of candidate solutions in terms of individual objectives, the algorithms can deliver a set of solutions satisfying as many objectives as possible. 

In software testing, many-objective search has been applied to solve testing problems with many test requirements.
The idea is to recast such testing problems into optimization problems by carefully defining fitness functions for all objectives.
However, since the number of objectives (i.e., test requirements) is often much more than four (e.g., covering \emph{all} branches in a program), researchers have developed tailored algorithms for testing. 
MOSA~\cite{mosa} is a well-known algorithm dedicated to test suite generation with many objectives. 
It uses an evolutionary algorithm to achieve each objective individually by effectively searching for uncovered objectives and keeping an archive for the best test cases achieving objectives.
By defining fitness functions to measure the likelihood of covering individual branches in a program, MOSA can effectively and efficiently generate a test suite covering as many branches as possible~\cite{mosa}. 
FITEST~\cite{fitest} is another state-of-the-art algorithm that extends MOSA to decrease the population size as the number of uncovered objectives decreases to improve efficiency, in contrast to MOSA that maintains the same population size during the entire search.

\section{Problem Description}\label{sec:problem}

In this section, we provide a precise problem description regarding the automated online testing of DNN-enabled systems (DES) by dynamically changing their environment during simulation. 
As a working example, we use a DNN-enabled autonomous driving system (DADS) to illustrate our main points, but the description can be generalized to any DES.

In online testing, a DADS under test is embedded into and interacts with its driving environment.
However, because of the risks and costs it entails, online testing is usually performed with a simulator rather than on-road vehicle testing. 
Using a simulator enables the control of the driving environment, such as weather conditions, lighting conditions, and the behavior of other actors (e.g., vehicles on the road and pedestrians).
During simulation, the DADS continuously interacts with the environment by observing the environment via the ego vehicle's sensors (e.g., camera and LIDAR) and driving the ego vehicle through commands (e.g., steering, throttle, and braking).
Due to the closed-loop interaction between the DADS and its environment, simulation is an effective instrument to check if any requirements violation can occur under realistic conditions. 
Notice that such violations can be triggered by dynamically changing the driving environment during simulation; for example, certain changes to the speed of the vehicle in front, which can often occur in practice due to impaired driving or sudden stops, can cause a collision.
The goal of DADS online testing, based on dynamically changing the environment, is to find a minimal set of test cases, each of them changing the environment in a different way, to cause the DADS to violate as many requirements as possible.

Specifically, let $d$ be the DADS under test on board the ego vehicle and $E = (X, K)$ be the environment where $X=\{x_1, x_2, \dots \}$ is a set of \emph{dynamic} elements of the simulation (e.g., actors other than the ego vehicle, and the environment conditions) and $K$ is a set of \emph{static} elements (e.g., roads, buildings, and trees). 
Each dynamic element $x\in X$ can be further decomposed into a sequence $x= \langle x^1, x^2, \dots, x^J \rangle$ where $J$ is the duration of the simulation and $x^j$ is the value of $x$ at time $j\in \{1,2,\dots,J\}$ (e.g., the position, speed, and acceleration of the vehicle in front at time $j$).
Based on that, we can define $X^j = \{ x^j \mid x\in X\}$ as capturing the set of values of all the dynamic elements in $X$ at time $j$ and $E^j = (X^j, K)$ as indicating the set of values of all environmental elements (both dynamic and static) in $E$ at time $j$.
Let a \emph{test case} $t = \langle X^1, \dots, X^J \rangle$ be a sequence of values of the dynamic elements for $J$ time steps. 
By running $d$ in $E$ with $t$, using a simulator, at each time step $j\in \{1,2,\dots,J\}$, $d$ observes the snapshot of $E^j = (X^j, K)$ using the sensors (e.g., camera and LIDAR) and generates driving commands $C^j$ (e.g., throttle, steering, and braking) for the ego vehicle. 
Note that what $d$ observes from the same $E$ varies depending on the ego vehicle's dynamics (e.g., position, speed, and acceleration) computed by $C$; for example, $d$ observes the \textit{relative} distance between the ego vehicle and the vehicle in front, which naturally varies depending on the position of the ego vehicle.
For the next time step $j+1$, the simulator computes $E^{j+1} = (X^{j+1}, K)$ according to $t$ and generates what $d$ observes from $E^{j+1}$ by taking into account $C^j$.

Given a set of requirements (safety, functional) $R$, the degree of a violation for a requirement $r\in R$ produced by $d$ in $E$ for $t$ at any time step $j\in \{1,\dots,J\}$, denoted by $v(r, d, E, t, j)$, can be measured by monitoring the simulation of $d$ in $E$ for $t$. 
For example, the distance between the ego vehicle and the vehicle in front can be used to measure how close they are from colliding. 
If $v(r, d, E, t, j)$ is greater than a certain threshold $\epsilon_r$ (i.e., the distance is closer than the minimum safe distance) at any $j$, we say that $d$ \textit{violates} $r$ at $j$.
Let $v_\mathit{max}(r, d, E, t)$ be the maximum degree of violation for $r$ produced by $d$ in $E$ for $t$ during simulation. 
Given an initial environment $E^1 = (X^1, K)$, the problem of DADS online testing for dynamically changing environments is to find a minimal set of test cases $\mathit{TS}$ that satisfies $v_\mathit{max}(r, d, E, t) > \epsilon_r$ for as many $r\in R$ as possible when executing all $t\in \mathit{TS}$. 

Test data generation for online testing of a DADS, with a dynamically changing environment, presents several challenges. 
First, the input space for dynamically changing the behavior of the environment is enormous because there are many possible combinations of environmental changes for each timestamp.
Second, there are usually many independent requirements to be considered simultaneously. 
For example, keeping a safe distance from the vehicle in front is independent from the ego vehicle abiding by the traffic lights. 
If the requirements are not considered simultaneously, no practical test budget may be sufficient to thoroughly test each requirement, as a limited budget must be divided across all individual requirements.
Third, in addition to the second challenge, depending upon the accuracy of the DADS under test, it may be infeasible to violate some requirements.
This implies that the pre-assigned budgets for those requirements are inevitably wasted if a testing approach cannot simultaneously consider all requirements.
Last but not least, a DADS is often developed by a third party, and as a result its internal information (e.g., about DNN models) is often not fully accessible. Therefore, online DADS testing must often be carried out in a black-box manner.

To address the challenges mentioned above, we propose a novel approach that combines two distinct approaches: 
\begin{inparaenum}[(1)]
\item RL to dynamically change the environment based on the simulation state (including the state of the DES under test and the state of the environment) at each timestamp with the aim of causing requirements violations, and
\item many-objective search to effectively and efficiently achieve many independent objectives (i.e., violating requirements in our context).
\end{inparaenum}
Furthermore, our approach is DNN-agnostic; it does not need any internal information about the DNN. 

\section{Related Work}\label{sec:related-work}

This section discusses DES (DNN-Enabled Systems) testing in terms of two approaches closely related to ours: search-based testing and RL-based testing.

\subsection{Search-Based Testing}\label{sec:related-work:search}
Search-based testing has been widely adapted to test DES, particularly DADS, by formulating a test data generation problem as an optimization problem where an objective is to cause the DADS under test to misbehave and applying metaheuristic search algorithms to solve the optimization problem automatically.
For example, \citet{asfault} presented \textsc{AsFault}, a tool for generating test scenarios in the form of road networks using a genetic algorithm to cause the DADS under test to go out of lane.
\citet{8814230} tested motion-planning modules in DADS by generating critical test scenarios based on a minimization of the search space, defined as a set of scenario parameter intervals, of the vehicle under test.
\citet{riccio2020model} presented \textsc{DeepJanus}, an approach that uses NSGA-II to generate a \textit{pair} of close scenarios, where the DADS under test misbehaves for one scenario but not for the other, in an attempt to find a \textit{frontier} behavior at which the system under test starts to misbehave. 
For a configurable, parameterized ADS, \citet{9159061} aimed to find \textit{avoidable} collisions, that is collisions that would not have occurred with differently-configured ADS; using NSGA-II, they first search for a collision scenario and then search for a new configuration of the ADS which avoids the collision.
Considering the high computational cost of simulations involved in search-based approaches, improving the efficiency of search-based testing has also been studied in the context of DES/DADS testing.
\citet{abdessalem2018testing} presented \textsc{NSGAII-DT} that combines NSGA-II (a multi-objective search algorithm) and Decision Tree (a classification model) to generate critical test cases while refining and focusing on the regions of a test scenario space that are likely to contain cost critical test scenarios. 
\citet{ul2022efficient} presented \textsc{SAMOTA}, an efficient online testing approach extending many-objective search algorithms tailored for test suite generation to utilize \textit{surrogate models} that can mimic driving simulation (e.g., whether a requirement violation occurred or not) and are much less expensive to run. 
\citet{li2020av} presented \textsc{AVFuzzer}, a single-objective approach to find safety violations in autonomous vehicles by changing driving manoeuvres of other vehicles on the road in a coarse-grained manner (e.g., lane changes).

Though search-based testing approaches have shown to work very well for testing DADS when the environment remains static throughout the tested scenario, they are not likely to work efficiently in the case of dynamically changing environments. 
One of the reasons is they must get the fitness score for a test scenario after the simulation is completed (when we can ascertain whether a requirement violation occurred) and cannot, therefore, change the environment during a simulation run. 
Another challenge is that the search space becomes enormous because of the many possible combinations of environment parameters at each timestamp.

\subsection{RL-Based Testing}\label{sec:related-work:rl}
RL has also recently been used for DES/DADS testing.
In RL-based testing, a DES/DADS testing problem is formulated as a sequential decision-making problem where the goal is to generate a compact test scenario (in the form of a sequence of environmental changes) that causes the system under test to violate a given requirement. 
For example, \citet{koren2018adaptive} presented an approach that extends Adaptive Stress Testing (AST)~\cite{lee2015adaptive} by using reinforcement learning for updating the environment of a vehicle to cause a collision.
\citet{corso2019adaptive} further extended AST to include domain relevant information in the search process. This modification helps in finding a more diverse set of test scenarios in the context of DADS testing.
\citet{sharif2021adversarial} presented a two-step approach that not only generates test scenarios using Deep RL but also utilizes the test scenarios to improve the robustness of the DADS under test by retraining it.
Very recently, \citet{lu2022learning} presented \textsc{DeepCollision}, an approach that learns the configurations of the environment, using Deep Q-learning, that can cause the crash of the ego vehicle. 

Despite encouraging achievements  when using RL for DES/DADS testing, especially when the objective is to dynamically change the environment, there is no work that focuses on testing many independent requirements simultaneously, thus raising the problems discussed in Section~\ref{sec:problem}. 

\section{Reinforcement Learning-Based Test Generation}\label{sec:approach}
This section presents MORLOT (Many-Objective Reinforcement Learning for Online Testing), our novel approach to address the problem explained in Section~\ref{sec:problem}. In the following subsections, we first describe how Reinforcement Learning (RL) can be tailored for the generation of a single test case (i.e., a test scenario in the context of DES online testing), and then present MORLOT by extending it.

\subsection{Test Case Generation using RL}\label{sec:approach:single}
RL has widely been used to learn the sequence for completing a sequential decision-making task~\cite{sutton2018reinforcement,aar6404}. 
RL has also been applied to automated software testing~\cite{zheng2021automatic,zheng2019wuji,pan2020reinforcement}.
For the latter, RL is particularly suitable for systems whose usage entails sequential steps, for example ordering something from the web~\cite{zheng2021automatic} such as: (1) going to the website, (2) putting something in the cart, (3) checkout and payment. 
Similarly, in the case of testing a DADS, we require sequential changes in the environment; for example, sequential steps for one scenario can be: (1) change the weather to \textit{Rainy} (to decrease the friction between tyres and the road), (2) increase the \textit{fog} level (to reduce visibility), (3) increase the speed of the vehicle-in-front (to increase the distance from the ego-vehicle, which then speeds up as no obstacle is visible), (4) abruptly slow the vehicle-in-front to trigger a collision (violation of safety requirement).

To generate a test case (i.e., a test sequence) for a single requirement (safety, functional), RL is driven by an objective that must be satisfied while interacting with the environment. 
The objective is to find any test case $t$ that satisfies $v_\mathit{max}(r, d, E, t)>\epsilon_r$, where $v_\mathit{max}(r, d, E, t)$ is the maximum degree of violation for a requirement $r$ observed over the simulation of the DADS under test $d$ in its driving environment $E$, while executing $t$ and assuming $\epsilon_r$ is the threshold specifying the maximum acceptable violation for $r$. 
The goal of RL-based testing is to find a sequence of changes in the environment that results in satisfying the objective; these changes are stored in $t$ in the form of state-action pairs, where a state captures a snapshot of $E$ and the ego vehicle and an action indicates the change to be applied to $E$ given the state. 
Storing states in $t$ is essential as it provides necessary information for explaining the changes in the environment that resulted in a requirement violation.
A single test case is therefore composed of a sequence of state-action pairs leading to requirement violations. 

As described in Section~\ref{sec:background}, RL methods can be categorised into two types: (1) tabular-based and (2) approximation-based.
Considering the simplicity and fast convergence of tabular-based methods, we use Q-learning~\cite{watkins1992q}, one of the most widely used algorithms in this category, as our basis in the rest of the paper. Nevertheless, one can easily opt for other tabular-based RL methods, such as SARSA~\cite{rummery1994line}, by just changing the way of updating the Q-table. 

To use Q-learning for testing, it is essential to define states, actions, and rewards for an RL agent as described in Section~\ref{sec:background:rl}.
In the context of DADS testing, \textit{states} can be defined to capture important details of the simulation (e.g., locations/speeds of actors, weather conditions), \textit{actions} are the environment changes (e.g., change in the dynamics of actors and weather conditions) and \textit{rewards} should indicate the degree of requirements violations. The higher the degree of violation, the higher the reward, so that the RL agent can generate a sequence of state-action pairs that maximizes the sum of rewards.

Algorithm~\ref{alg:mo} presents a generic RL-based testing algorithm that takes as inputs an objective $o$, an environment $E$ and a Q-table $q$ (possibly initialized based on prior knowledge), and returns a test case $t$ achieving $o$ and a Q-table $q$ that was updated during the generation of $t$. 
If the algorithm cannot find a $t$ that satisfies $o$, it returns a null value for $t$ along with the updated Q-table $q$ resulting from the search, which can be reused later if needed. 

\begin{algorithm}
\SetKwInOut{Input}{Input}
\SetKwInOut{Output}{Output}

\Input{Objective $o$,\\
    Environment $E$,\\
    Q-table $q$}
\Output{Test Case $t$}
     \While{$\mathit{not (\mathit{budget\_finished})}$}{
        Test Case $t \gets \emptyset$ \label{alg:qt:tc} \\
        $E \gets \mathit{reset(E)}$ \label{alg:qt:init}\\
        \While{$\mathit{not (\mathit{stopping\_condition})}$\label{alg:qt:loop-start}}{
        State $s \gets \mathit{observe}(E)$ \label{alg:qt:env} \\
        Action $a \gets \mathit{chooseAction}(q,s)$ \label{alg:qt:acc}\\
        Reward $w \gets \mathit{perform}(E,a)$\label{alg:qt:atc}\\
        $t \gets \mathit{append} (t,(s,a))$ \label{alg:qt:append}\\
        $q \gets \mathit{updateQtable}(q,s,a,w)$\label{alg:qt:update}\\
        \If{$\mathit{satisfy}(t,o)$}
        {
            \textbf{return} $t$, $q$\\
        }
\label{alg:qt:loop-end}
        }
    }
    \textbf{return} \textbf{$\mathit{null,q}$}
\caption{RL-based Test Generation (single objective)}
\label{alg:mo}
\end{algorithm}

The algorithm begins with the loop for finding $t$ that satisfies $o$.
Until the budget (e.g., total number of hours or simulator runs) runs out, 
the algorithm repeats the following steps: 
(1) initialize $t$ and resetting $E$ to its initial state (lines~\ref{alg:qt:tc}--\ref{alg:qt:init}) and
(2) run the tabular-based RL algorithm to generate $t$ (i.e., a sequence of environmental changes in the form of state-action pairs) with the aim of satisfying $o$ (lines~\ref{alg:qt:loop-start}--\ref{alg:qt:loop-end}; see below).
The algorithm ends by returning $t$ if $o$ is satisfied; otherwise, a null value is returned for $t$. 

To generate $t$ so that it satisfies $o$ (lines~\ref{alg:qt:loop-start}--\ref{alg:qt:loop-end}), the algorithm repeats the following steps until the stopping condition (e.g., satisfying $o$ or no more possible actions) is met:
\begin{inparaenum}[(i)]
    \item observe the state $s$ from $E$ (line~\ref{alg:qt:env}), 
    \item choose an action $a$ either randomly (with a small probability $\epsilon$ to increase the exploration of the state space and to avoid being stuck in local optima) or using $q$ and $s$ (line~\ref{alg:qt:acc}),
    \item perform $a$ to update $E$ and receive a reward $w$ for $o$ (line~\ref{alg:qt:atc}), 
    \item append a new state-action pair $(s,a)$ at the end of $t$ (line~\ref{alg:qt:append}), 
    \item update $q$ using $s$, $a$ and $w$ 
    (line~\ref{alg:qt:update}), and 
    \item return $t$ and $q$ if the objective $o$ is satisfied (i.e., a violation is found) (line~\ref{alg:qt:loop-end}).
\end{inparaenum}

\subsection{Test Suite Generation using many-objective RL}\label{sec:approach:many}
Algorithm~\ref{alg:mo} works well with one objective (violating one requirement) while the nature of our problem, as described in Section~\ref{sec:problem}, involves multiple independent objectives. Therefore, we need to extend the algorithm above to efficiently take into account many objectives.

As discussed in Section~\ref{sec:related-work:search}, there is existing work on covering many independent objectives in the context of DES/DADS testing~\cite{abdessalem2018testing,fitest,ul2022efficient}.
Though they test both static and dynamic elements of the environment, they do not change the dynamic elements during the execution (simulation) of a test case.
Existing approaches can be used for the problem of DADS testing with dynamically changing environments if they extend the search space to take into account the environment's dynamic elements over a certain time horizon; however, this would be highly inefficient due to the resulting much larger search spaces (see Section~\ref{sec:eval} for details).

To efficiently solve the problem of DES online testing considering dynamically changing environments, with many independent objectives, we propose a novel approach: Many-Objective Reinforcement Learning for Online Testing (MORLOT). It combines two distinct techniques: 
(1) \emph{tabular-based Reinforcement Learning (RL)} to dynamically interact with the environment for finding the environmental changes that cause the violation of given requirements and
(2) \emph{many-objective search} for test suite generation~\cite{mosa,fitest,ul2022efficient} to achieve many independent objectives (i.e., violating the requirements) individually within a limited time budget.

Similar to existing work, MORLOT uses the notion of archive to keep the minimal set of test cases satisfying the objectives.
To take into account many independent objectives simultaneously, we extend Algorithm~\ref{alg:mo} to have multiple Q-tables, each of them addressing one objective. 
Intuitively, each Q-table captures the best action to select for one corresponding objective in a given state. 
However, the challenge is that, in the same states, different actions can be chosen for different objectives (by different Q-tables).
To choose a single action to perform, we select the Q-table based on the objective that achieved the maximum fitness value (i.e., reward in RL) in the previous iteration. This is because that objective is the closest to being satisfied.

MORLOT takes a set of objectives $O$, an environment $E$ and a set of Q-tables $Q$ (possibly initialized based on prior knowledge); MORLOT returns a test suite containing a test case for each satisfied objective.
As stated earlier, we define each objective as a violation of a certain requirement. 
Specifically, given a set of requirements $R=\langle r_1, r_2, \dots r_n \rangle$ for the DADS $d$, we define a set of objectives $O=\langle o_1, o_2, \dots o_n\rangle$ where $o_i$ is to cause $d$ to violate $r_i$ for $i=1,2,\dots,n$.
MORLOT returns a test suite $TS = \langle t_1, t_2, \dots, t_m\rangle$ where $t_l$ is a test case satisfying any one of the objectives $o_i \in O$ and $m \leq n$.

Algorithm~\ref{alg:rl-mo} shows the pseudocode of MORLOT. It takes $O$, $E$ and $Q$ as inputs and returns a test suite containing test cases satisfying at least one objective and multiple Q-tables, one for each objective.

\begin{algorithm}
\SetKwInOut{Input}{Input}
\SetKwInOut{Output}{Output}

\Input{ {Set of Objectives $O$}\\
    Environment $E$\\
    Set of Q-tables $Q$
    }
\Output{{Archive (Test Suite) $A$\\ Set of Q-tables $Q$}}

{ Set of Uncovered Objectives $U \gets O$\label{alg:mo-rl:uc}\\}
  
    Archive $A \gets \emptyset$ \label{alg:mo-rl:arc}\\
    \While{$\mathit{not (\mathit{budget\_finished})}$}{
        Set of Rewards $W \gets \emptyset$\label{alg:mo-rl:init-R}\\
        Test Case $t \gets \emptyset$\label{alg:mo-rl:init-t}\\
        $E \gets \mathit{reset(E)}$ \label{alg:mo-rl:reset} \\
        
        \While{$\mathit{not (\mathit{stopping\_condition})}$}{\label{alg:mo-rl:while:start}
            State $s \gets \mathit{observe}(E)$ \label{alg:mo-rl:observe}\\
            Action $a \gets \mathit{chooseActionMultiObjs}(s,{R,Q,U})$\label{alg:mo-rl:chooseAcc}\\
            $W \gets \mathit{performMultiObjs}(a,E)$ \label{alg:mo-rl:setOfR}\\
            $Q \gets \mathit{{updateQtables}(Q,s,a,W)}$ \label{alg:mo-rl:updateQ}\\
            $t \gets \mathit{append} (t,(s,a))$\label{alg:mo-rl:update-t}\\
            \ForEach{$o \in O$\label{alg:mo-rl:updateArch}}{
                \If{$\mathit{satisfy}(t,o)$}{
                    $A \gets \mathit{updateArchive}(A,t,o)$ \\
                    $U \gets U - \{o\}$\\ \label{alg:mo-rl:update-u}
                }
            }
        }\label{alg:mo-rl:while:end}
    }
    \textbf{return} \textbf{$\mathit{{A, Q}}$}
\caption{MORLOT}
\label{alg:rl-mo}
\end{algorithm}

The algorithm starts by initializing the set of uncovered objectives $U$ with $O$ (line~\ref{alg:mo-rl:uc}). It is important to keep a record of uncovered objectives so that the search process can focus on them.
It then initializes $A$ (line~\ref{alg:mo-rl:arc}). Notice that $|Q| = |O|$ so that there is a Q-table for each objective. 
Until the search budget runs out, the algorithm repeats the following steps:
(1) initialize a set of rewards $W$ and a test case $t$ and resetting $E$ to its initial state (lines~\ref{alg:mo-rl:init-R}--\ref{alg:mo-rl:reset}) and
(2) find $t$ that satisfies $u \in U$ using RL (lines~\ref{alg:mo-rl:while:start}--\ref{alg:mo-rl:while:end}).
To achieve the latter, the algorithm repeats the following steps until the stopping conditions are met:
\begin{inparaenum}[(i)]
    \item observe $s$ from $E$ (line~\ref{alg:mo-rl:observe}), 
    \item choose an action $a$ either randomly (with a small probability $\epsilon$ to increase the exploration of the state space and to avoid being stuck in local optima) or using a Q-table $q_m\in Q$ and $s$ where $q_m$ is the Q-table of an uncovered objective $u\in U$ whose reward $w\in W$ for the previously chosen action is the maximum  (line~\ref{alg:mo-rl:chooseAcc}),
    \item perform $a$ to update $W$ received from $E$ (line~\ref{alg:mo-rl:setOfR}),
    \item update $Q$ using $s$, $a$, and $W$ (line~\ref{alg:mo-rl:updateQ}),
    \item append $(s, a)$ at the end of $t$ (line~\ref{alg:mo-rl:update-t}), and
    \item update $A$ and $U$, if $t$ satisfies any $o_i\in O$, such that $A$ includes the shortest test case satisfying $o_i$ from $A \cup \{t\}$ and $U$ excludes $o_i$  (lines~\ref{alg:mo-rl:updateArch}--\ref{alg:mo-rl:update-u}).
\end{inparaenum}
The algorithm ends by returning $A$ (i.e., a minimal set of test cases, each of them covering at least one objective) and $Q$ (i.e., a set of filled Q-tables, each of them matching one objective).

Notice that MORLOT updates the Q-tables $Q$ even for covered objectives, while addressing the uncovered objectives, as $Q$ can be reused later for a newer version of the DES under test in a regression testing setting. Since the Q-tables record the best actions to choose for given states, using them for testing the newer versions of the DES can boost the performance of Algorithm~\ref{alg:mo}. This investigation is however left to future work.

\section{Evaluation}\label{sec:eval}
This section reports on the empirical evaluation of MORLOT when testing an open-source DADS. Specifically, we answer the following research questions:
\begin{enumerate}[\bf RQ1:]
    \item How does MORLOT fare compared to other many-objective search approaches tailored for test suite generation in terms of \textit{test effectiveness}?
    \item How does MORLOT fare compared to other many-objective search approaches tailored for test suite generation in terms of \textit{test efficiency}?
\end{enumerate}

To answer RQ1, we compare test suites generated by different approaches within the same execution time budget (in computing hours) in terms of their ability to reveal safety and functional requirements violations. 
To answer RQ2, we compare different approaches in terms of the execution time required to reveal a certain number of requirements violations and how differences among them evolve over time. 
These investigations aim to evaluate the benefits of MORLOT for DADS online testing, in terms of test effectiveness and efficiency, and therefore the benefits of dynamically changing the environment based on the simulation state.

\subsection{Evaluation Subjects}\label{sec:eval:subject}

We use TransFuser (TF)~\cite{Prakash2021CVPR}, the highest rank DADS among publicly available ones in the CARLA Autonomous Driving Leaderboard Sensors Track~\cite{carla-leaderboard} at the time of our evaluation.
The Leaderboard evaluates the driving performance of ADS in terms of 11 different metrics designed to assess driving safety, such as red light infractions, collision infractions, and route completion.
The driving performance results of TF reported in the Leaderboard show that it is well-trained and able to pass a large variety test scenarios. It ought therefore to be representative of what one can find in the industry.

TF takes an image from the front-facing camera and the sensor data from LiDAR as input and generates the driving command (steering, throttle, and braking). 
Internally, it uses ResNet34 and ResNet18~\cite{he2016deep} to extract features from the input image and sensor data, respectively.
It then uses transformers~\cite{vaswani2017attention} to integrate the extracted image and LiDAR features.
The integrated features are processed by a way-point prediction network that predicts the ego vehicle's expected trajectory, which is used for determining the driving command for next time steps.

We also use CARLA~\cite{CARLA-citation}, a high-fidelity open-source simulator developed for autonomous driving research. 
CARLA provides hand-crafted static and dynamic elements for driving simulations. 
Static elements include different types of roads, buildings, and traffic signs. 
Dynamic elements include other vehicles, pedestrians, weather, and lighting conditions.
In our evaluation, we let the approach under evaluation (i.e., MORLOT and its alternatives) control a subset of dynamic elements to mimic real-world scenarios, such as weather and lighting conditions and the behavior of pedestrians, which are dynamically controllable during the simulation. 
They also control the throttle and steering of the Vehicle-In-Front (VIF), which is one of the most influential factors in the driving performance of the Ego Vehicle (EV).
Furthermore, to avoid trivial violations of safety and functional requirements resulting from the behavior of dynamic elements (e.g., a pedestrian runs into the EV), we manually imposed constraints on such behaviors.
The details of the constraints can be found in the supporting material (see Section~\ref{sec:data}).

Considering the capability of the simulator, we use the following six safety and functional requirements: 
\begin{enumerate}[$r_1$:]
\item the EV should not go out of lane;
\item the EV should not collide with other vehicles; 
\item the EV should not collide with pedestrians;
\item the EV should not collide with static meshes (i.e., traffic lights, traffic signs etc.);
\item the EV should reach its destination in defined time budget;
\item the EV should not violate traffic lights.
\end{enumerate}

Recall that we should specify an initial environment that determines the static elements and the initial states of the dynamic elements for simulation. 
In practice, one can randomize the initial environments to test diverse scenarios. 
In our evaluation, however, we need the same initial environments for different approaches (and their repeated runs) to compare them fairly in terms of test effectiveness and efficiency.
Since the road type defined in the initial environment is one of the critical factors that has the greatest influence on the driving performance of a DADS, we consider three different initial environments having three different road types: \textit{Straight}, \textit{Left-Turn}, and \textit{Right-Turn}. 
We select \textit{Straight}, \textit{Left-Turn}, and \textit{Right-Turn} roads from Town05, one of the default maps provided in CARLA, as all the other maps were used for training Transfuser~\cite{Prakash2021CVPR}.
For the other environmental elements, we use the basic configuration (i.e., sunny weather, the VIF is 10 meters away from the EV, pedestrians are 20 meters away from the EV on a footpath, zero precipitation deposit on roads) provided in CARLA~\cite{CARLA-citation}. 
The details of the initial environment setup can be found in the supporting material (see Section~\ref{sec:data}).

Due to the execution time of individual simulations in CARLA (i.e., 5 minutes on average), the total computing time for all the three different initial environments is more than 600 hours (25 days).
To address this issue, we conduct our evaluation on two platforms, P1 and P2. 
Platform P1 is a desktop with Intel i9-9900K CPU, RTX 2080 Ti (11 GB) GPU, and 32 GB memory, running Ubuntu 18.04.
Platform P2 is a \texttt{g4dn.xlarge} node configured as Deep Learning AMI (version 61.1) in Amazon Elastic Cloud (\url{https://aws.amazon.com/ec2/}) with four virtual cores, NVIDIA T4 GPU (16GB), and 16 GB memory, running Ubuntu 18.04. 
Specifically, we use P1 for the \textit{Straight} environment and five instances of P2 for the remaining. 
By doing this, we can compare the results of different approaches (i.e., MORLOT and its alternatives) for the same initial environment.

\subsection{RQ1: Test Effectiveness}\label{sec:eval:effectiveneses}

\subsubsection{Setup}\label{sec:eval:eff:setup}
To answer RQ1, we generate test suites using MORLOT and other many-objective search approaches tailored for test suite generation using the same execution time budget. 
We compare the approaches in terms of \textit{Test Suite Effectiveness (TSE)}. 
Specifically, the \textit{TSE} of a test suite $TS$ is defined as the proportion of requirements $TS$ violated over the total number of requirements (i.e., six as explained in Section~\ref{sec:eval:subject}). 

To use MORLOT, we need to define states, actions, and rewards specific to our case study as we rely on a tabular-based RL method as mentioned in Section~\ref{sec:approach:single}.

In the context of DADS online testing, a state should contain all the information that may affect the requirements violations of the DADS, such as weather conditions and the dynamics of the Ego Vehicle (EV), Vehicle-In-Front (VIF), and pedestrian in terms of positions, speeds, and accelerations.
To reduce the state space, we consider only one VIF and one pedestrian since they are sufficient to generate critical test scenarios. Further, we rely on the spatial grid~\cite{leurent2018survey} and divide the road into 10x10 grids when representing the positions of the EV and VIF. 
For speed and acceleration, we use values reported by CARLA, rounded to one decimal point.
Specifically, we define a state $s$ as a 6-tuple $s=(\mathit{EV}, \mathit{VIF}, P, h, f, g)$ where each of its elements is defined as follows:
\begin{itemize}[-]
\item $\mathit{EV} = (x^\mathit{EV}, y^\mathit{EV}, v^\mathit{EV}_x, v^\mathit{EV}_y, a^\mathit{EV}_x, a^\mathit{EV}_y)$ is the state of the EV where $x^\mathit{EV}$ and $y^\mathit{EV}$ are the $x$ and $y$ components of the absolute position of the EV on the road, $v^\mathit{EV}_x$ and $v^\mathit{EV}_y$ are the $x$ and $y$ components of the absolute speed of the EV, and $a^\mathit{EV}_x$ and $a^\mathit{EV}_y$ are the $x$ and $y$ components of the absolute acceleration of the EV.
\item $\mathit{VIF} = (x^\mathit{VIF}, y^\mathit{VIF}, v^\mathit{VIF}_x, v^\mathit{VIF}_y, a^\mathit{VIF}_x, a^\mathit{VIF}_y)$ is the state of the VIF where $x^\mathit{VIF}$ and $y^\mathit{VIF}$ are the $x$ and $y$ components of the position of the VIF relative to the EV, $v^\mathit{VIF}_x$ and $v^\mathit{VIF}_y$ are the $x$ and $y$ components of the speed of the VIF relative to the EV, and $a^\mathit{VIF}_x$ and $a^\mathit{VIF}_y$ are the $x$ and $y$ components of the acceleration of the VIF relative to the EV.
\item $P = (x^P, y^P, v^P_x, v^P_y)$ is the state of the pedestrian where $x^P$ and $y^P$ are the $x$ and $y$ components of the direction of the pedestrian and $v^P_x$ and $v^P_y$ are the $x$ and $y$ components of the speed of the pedestrian. We do not consider the acceleration of the pedestrian since it is not computed in CARLA. 
\item $h$ is the weather state and can have a value ranging between 0 (clear weather) and 100 (thunderstorm) in steps of 2.5.
\item $f$ is the fog state and can have a value ranging between 0 (no fog) and 100 (heavy fog) in steps of 2.5.
\item $g$ is the lighting state (controlled by changing the location of the light source) and can have a value ranging between -30 (night) and 120 (evening) in steps of 2.5.
\end{itemize}

For actions, we keep the size of unit changes of the dynamic elements small, based on preliminary experiments, to avoid unrealistic changes within each time step; for example, the VIF's throttle can increase/decrease by only 0.1 at each simulation time step. As a result, for each time step, one of the following actions can be taken:
\begin{itemize}[-]
    \item increasing/decreasing throttle by 0.1 (throttle range: 0--1), 
    \item increasing/decreasing steering 0.01 (steering range: -1--1),
    \item increasing/decreasing light intensity by moving the source of light by 2.5 degrees,
    \item increasing/decreasing weather intensity by 2.5,
    \item increasing/decreasing fog intensity by 2.5,
    \item increasing/decreasing pedestrian speed by 0.05 m/s (speed range 0.3--1.5),
    \item changing pedestrian direction (x,y-axis) by 0.1 (direction change range: -1--1), and
    \item do nothing.
\end{itemize}

For rewards, we define one reward function for each requirement $r_i$ for $i=1,\dots,6$, discussed in Section~\ref{sec:eval:subject}. 
Since we aim to cause the DADS to violate the requirements, we need higher reward values for more critical situations. For $i=1,\dots,5$, the criticality of a situation depends on the distance between the EV and the object (i.e., the VIF, pedestrian, static meshes, center of the lane, and destination for $r_1$, $r_2$, $r_3$, $r_4$, and $r_5$, respectively); the shorter the distance, the more critical. 
To capture this, we define the reward function $\mathit{reward}_{1,\dots,5}$ for $r_1,\dots,r_5$ as follows:
\begin{equation*}
  \mathit{reward}_{1,\dots,5}=\begin{cases}
    1/d_{\mathit{EV}, \mathit{obj}}, & \text{if $d_{\mathit{EV}, \mathit{obj}} > 0$}\\
    1000000, & \text{else}
  \end{cases}
  \label{reward_equation}
\end{equation*}
where $d_{\mathit{EV}, \mathit{obj}}$ refers to the distance between the EV and the object and 1,000,000 is the maximum reward for any violation, a large number which is not achievable without violations.
The distance is normalized between 0 and 1 so that every requirement contributes equally to the reward function.
For $r_6$, $\mathit{reward}_6$ is defined as binary due to the limitation of CARLA: it returns 1 if $r_6$ is violated (i.e., at least one traffic rule is violated); otherwise 0.

Besides the definition of states, actions, and rewards for MORLOT, there are a few RL parameters to be tuned~\cite{sutton2018reinforcement}, such as the probability $\epsilon$ of choosing a random action, the learning rate $\alpha$, and the discount factor $\gamma$. 
For $\epsilon$, to make MORLOT more exploratory at first but gradually more exploitative, we dynamically decrease it from 1.0 to 0.1 during the search, following a suggested approach~\cite{mnih2015human}. 
Specifically, we decrease it from 1 to 0.1 in the first 20\% of the search budget and keep it at 0.1 for the rest.
For $\alpha$ and $\gamma$, based on preliminary experiments, we set the values to 0.01 and 0.9, respectively.

For comparison with MORLOT, we use two well-known approaches for many-objective test suite generation: MOSA~\cite{mosa} and FITEST~\cite{fitest}.
The six reward functions defined above for MORLOT are used as fitness functions for MOSA and FITEST.
Recall that, as described in Section~\ref{sec:problem}, our test case (test scenario) is a sequence of values of the dynamic elements for $J$ time steps, where the length of a simulation $J$ can vary depending on the behavior of dynamic elements. 
Since MOSA and FITEST search for test cases that satisfy many test objectives without sequentially determining environmental changes, the length of a test case (i.e., the length of a sequence of values of the dynamic elements) should be fixed before running the search. 
Therefore, we set the length of a test case as the maximum possible length of a simulation determined by the minimum car speed and the maximum road length; if a simulation ends before its maximum possible length, then the remaining sequence in a test case is ignored.
To be consistent with previous studies~\cite{ul2022efficient,fitest}, we set the population size equal to the number of objectives (i.e., requirements).
We follow the original studies~\cite{fitest,mosa} for mutation and crossover rates.

We also use Random Search (RS) with an archive as a baseline. 
RS generates random changes to the dynamic elements as a test case. 
The possible changes are the same as those of MORLOT, MOSA, and FITEST. 
Furthermore, RS maintains an archive for all the test cases that lead to requirements violations. 
RS will provide insights on how complex the search problem is and will help us quantify the relative effectiveness of advanced approaches: MORLOT, MOSA, and FITEST.

One might consider using MORL~\cite{yang2019generalized} or SAMOTA~\cite{ul2022efficient} as additional baselines. However, the former is a general multi-objective reinforcement learning approach that solves multiple competing objectives while MORLOT solves many independent objectives. Further, SAMOTA aims to reduce testing costs by using surrogate models that can predict the outputs of high-fidelity simulators, instead of running them, and thus assess at a much lower cost the degree of safety violations for each candidate test case during the search. Since a test case in our context is a sequence of actions with a length up to 2500, it is unrealistic to expect accurate surrogate models that take such a long sequence as input, and therefore, we could not reasonably compare MORLOT with SAMOTA.

Note that we do not additionally consider other RL-based testing approaches for comparison since they do not account for cases where many requirements must be validated at the same time.
As explained in Section~\ref{sec:problem}, simply repeating a single-objective approach multiple times, by dividing the test budget per requirement, cannot scale in our context, given that test executions are expensive due to high-fidelity simulations.

To account for randomness in all the approaches, we repeat the experiment 10 times with the same time budget of four hours. 
We found that fitness reaches a plateau after four hours based on our preliminary evaluations. 
We apply Mann–Whitney U tests~\cite{mann1947test} to evaluate the statistical significance of differences in \textit{TSE} values among the approaches. 
We also measure Vargha and Delaney's $\hat{A}_{AB}$~\cite{5002101} to calculate the effect size of the differences; $\hat{A}_{AB}$ ($= 1-\hat{A}_{BA}$) indicates that $A$ is better than $B$ with a small, medium, and large effect size when its value exceeds 0.56, 0.64, and 0.71, respectively.

\subsubsection{Results}

\begin{figure}
    \centering
	\includegraphics[width=\linewidth]{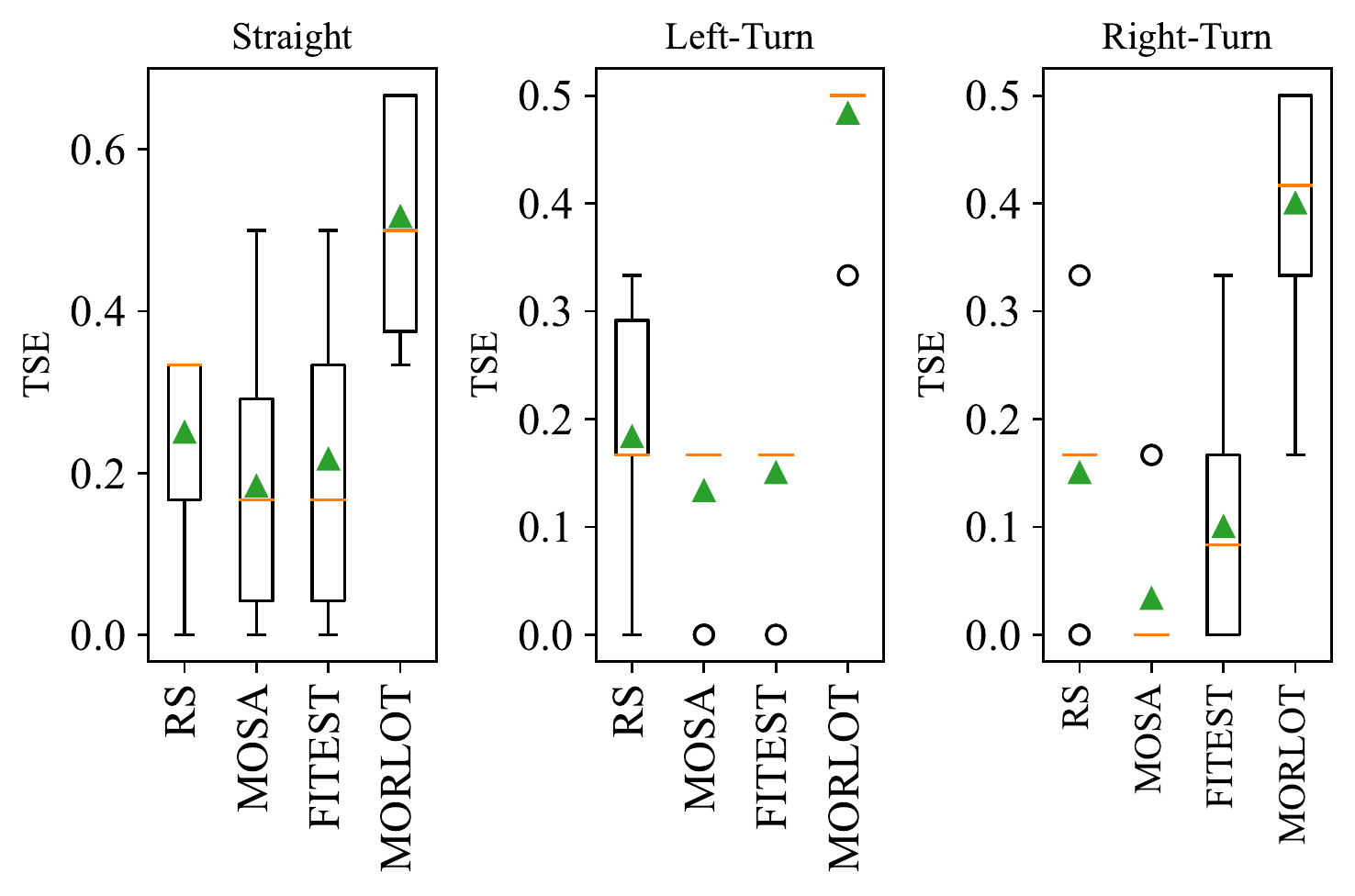}
	\caption{Distribution of \textit{TSE} values for different testing approaches}
	\label{fig:tse}
\end{figure}

\begin{table}
\small
\caption{Statistical comparison results for different approaches}
\label{table:tse-comparison}
\centering
\resizebox{\linewidth}{!}{%
\begin{filecontents*}{data/pvaluesRQ1-updated.csv}
A,B,Pvalues-config,Effectsize-config,Pvalues-config-2,Effectsize-config-2,Pvalues-config-3,Effectsize-config-3
FITEST,RS,0.605,0.43,0.2613,0.415,0.2694,0.365
MOSA,RS,0.2612,0.35,0.12,0.38,0.009294,0.19
MORLOT,RS,0.001,0.91,0,0.985,0.000638,0.935
MOSA,FITEST,0.7527,0.455,0.3939,0.45,0.1614,0.34
MORLOT,FITEST,0.003,0.88,0,1,0.0005207,0.95
MORLOT,MOSA,0.001,0.925,0,1,0.0001227,0.99
\end{filecontents*}
\pgfplotstabletypeset[
	col sep=comma,
	every head row/.style={before row={
        \toprule
        \multicolumn{2}{c}{\textit{Comparison}} &
        \multicolumn{2}{c}{\textit{Straight}} & 
        \multicolumn{2}{c}{\textit{Left-Turn}} & \multicolumn{2}{c}{\textit{Right-Turn}}\\
        \cmidrule(r){1-2} \cmidrule(r){3-4} \cmidrule(r){5-6} \cmidrule(r){7-8}
    }, after row=\midrule},
	every last row/.style={after row=\bottomrule},
    fixed,fixed zerofill,empty cells with={-},
    columns/A/.style={string type,column type=l,column name=A},
	columns/B/.style={string type,column type=l,column name=B},
	columns/Pvalues-config/.style={column type=r,column name=$p$-val,precision=3},
	columns/Effectsize-config/.style={column type=r,column name=$\hat{A}_{AB}$,precision=2},
	columns/A-2/.style={string type,column type=l,column name=A},
	columns/B-2/.style={string type,column type=l,column name=B},
	columns/Pvalues-config-2/.style={column type=r,column name=$p$-val,precision=3},
	columns/Effectsize-config-2/.style={column type=r,column name=$\hat{A}_{AB}$,precision=2},
	columns/A-3/.style={string type,column type=l,column name=A},
	columns/B-3/.style={string type,column type=l,column name=B},
	columns/Pvalues-config-3/.style={column type=r,column name=$p$-val,precision=3},
	columns/Effectsize-config-3/.style={column type=r,column name=$\hat{A}_{AB}$,precision=2},
]{data/pvaluesRQ1-updated.csv}}
\end{table}

Figure~\ref{fig:tse} shows the distribution of \textit{TSE} values achieved by RS, MOSA, FITEST, and MORLOT over 10 runs for each of the three initial environments (i.e., \textit{Straight}, \textit{Left-Turn}, and \textit{Right-Turn}). 
The orange bar and green triangle in the center of each box represent the median and average, respectively.
In addition, Table~\ref{table:tse-comparison} shows the statistical comparison results between different approaches. 
The columns $A$ and $B$ indicate the two approaches being compared. 
The columns $p$-value and $\hat{A}_\mathit{AB}$ indicate the statistical significance and effect size, respectively, when comparing $A$ and $B$ in terms of \textit{TSE}. 

For all three initial environments, it is clear that MORLOT outperforms RS, MOSA, and FITEST.
Given a significance level of $\alpha = 0.01$, the differences between MOTLOT and the others are significant ($p$-value $< 0.01$) in all cases.
Furthermore, $\mathit{\hat{A}_{AB}}$ is always greater than 0.71 when $A =$ MORLOT, meaning that MORLOT has a large effect size when compared to the others in terms of \textit{TSE}.
This result implies that, by combining RL and many-objective search, MORLOT can detect significantly more violations for a given set of safety and functional requirements than random search and state-of-the-art many-objective search approaches for test suite generation.
This is mainly because MORLOT can incrementally generate a sequence of changes by observing the state and reward after each change, whereas the other approaches must generate an entire sequence at once whose fitness score is calculated only after the simulation is completed.

It is also interesting to see that MOSA and FITEST do not outperform RS in all cases.
Given $\alpha=0.01$, the differences between MOSA and RS and between FITEST and RS are insignificant, except for the difference between MOSA and RS in the \textit{Right-Turn} environment.
This means that, in the \textit{Straight} and \textit{Left-Turn} environments, the advanced many-objective approaches are not significantly better than simple random search with an archive. 
One possible explanation can be that the search space considering dynamically changing environmental elements is enormously large, making advanced search approaches less effective within the given time budget (i.e., four hours).
Although they might perform better than random search if given much more time, this is unrealistic in practical conditions. 

\begin{table*}
\centering
\caption{Number of violations detected by different approaches}
\label{table:reqs-violation}
\begin{filecontents*}{data/reqs-table.csv}
Reqs,RS-1,MOSA-1,FITEST-1,MORLOT-1,RS-2,MOSA-2,FITEST-2,MORLOT-2,RS-3,MOSA-3,FITEST-3,MORLOT-3
r1,1,2,3,5,8,8,9,10,0,0,0,0
r2,0,0,0,10,0,0,0,10,0,0,0,2
r3,7,4,5,10,3,0,0,8,6,2,5,9
r4,0,0,0,0,0,0,0,0,0,0,0,0
r5,8,5,6,5,6,0,0,1,3,0,1,6
r6,0,0,0,1,0,0,0,0,0,0,0,7
\end{filecontents*}
\pgfplotstabletypeset[
	col sep=comma,
	every head row/.style={before row={
        \toprule
        \multicolumn{1}{c}{\textit{ }}&
        \multicolumn{4}{c}{\textit{Straight}}
        & 
        \multicolumn{4}{c}{\textit{Left-Turn}}& \multicolumn{4}{c}{\textit{Right-Turn}}\\
        \cmidrule(r){2-5} 
        \cmidrule(r){6-9}
        \cmidrule(r){10-13}
    }, after row=\midrule},
	every last row/.style={after row=\bottomrule},
    fixed,fixed zerofill,precision=0,empty cells with={-},
    columns/Reqs/.style={string type,column type=r,column name=Rq},
	columns/RS-1/.style={column type=r,column name=RS},
	columns/FITEST-1/.style={column type=r,column name=FITEST},
	columns/MOSA-1/.style={column type=r,column name=MOSA},
	columns/MORLOT-1/.style={column type=r,column name=MORLOT},
	columns/RS-2/.style={column type=r,column name=RS},
	columns/FITEST-2/.style={column type=r,column name=FITEST},
	columns/MOSA-2/.style={column type=r,column name=MOSA},
	columns/MORLOT-2/.style={column type=r,column name=MORLOT},
	columns/RS-3/.style={column type=r,column name=RS},
	columns/FITEST-3/.style={column type=r,column name=FITEST},
	columns/MOSA-3/.style={column type=r,column name=MOSA},
	columns/MORLOT-3/.style={column type=r,column name=MORLOT},
]{data/reqs-table.csv}
\end{table*}

To better understand which testing approaches detect violations of which requirements, Table~\ref{table:reqs-violation} shows the number of runs (among 10 repeats) that detect a violation for each requirement.
For example, a value of 1 in the first row and \textit{Straight}-RS column indicates that RS detects the violation of $r_1$ only once among the 10 repeats in the \textit{Straight} environment.
Requirement $r_4$ is never violated, meaning that the EV does not collide with static meshes (e.g., traffic signs) in any of the cases.
This is mainly because the static meshes are far enough from the road, making it difficult to make the EV collide with them. 
As for the remaining requirements, we can see that only MORLOT can detect the violations of $r_2$ and $r_6$.
A detailed analysis of the violations reveals that both $r_2$ and $r_6$ are highly relevant to the dynamics of the VIF; for example, MORLOT generated a VIF trajectory such that the VIF abruptly stops and slowly goes out of the camera frame, which causes the EV to collide.
This implies that only MORLOT can effectively change the dynamics of the VIF to cause the violations of $r_2$ and $r_6$ while the other approaches cannot. 
Note that there are no requirements that are not violated by MORLOT but are by another approach, further highlighting the test effectiveness of MORLOT. 

In Table~\ref{table:reqs-violation}, it is worth noticing that there are requirements for which MORLOT does not detect violations in all runs.
Though this could be partially improved by making MORLOT more exploratory (by increasing the $\epsilon$ value), it may affect the balance between exploration and exploitation.
Therefore, to better detect unknown violations in practice, it is recommended to run MORLOT multiple times if time permits.

\begin{tcolorbox}
The answer to RQ1 is that MORLOT is significantly more effective, in terms of \textit{Test Suite Effectiveness (TSE)}, and with a large effect size, than random search and alternative many-objective search approaches tailored for test suite generation.
\end{tcolorbox}

\subsection{RQ2: Test Efficiency}\label{sec:eval:efficiency}
\subsubsection{Setup}
To answer RQ2, we basically use the same setup as in RQ1 but additionally measure the achieved \textit{TSE} values at 20-minute intervals over a 4-hour run. To account for randomness, again, we repeat the experiment 10 times and report on how the average \textit{TSE} values for 10 runs vary over time from 20 minutes to 240 minutes in steps of 20 minutes.

\subsubsection{Results}

\begin{figure*}
    \centering
	\includegraphics[width=\linewidth]{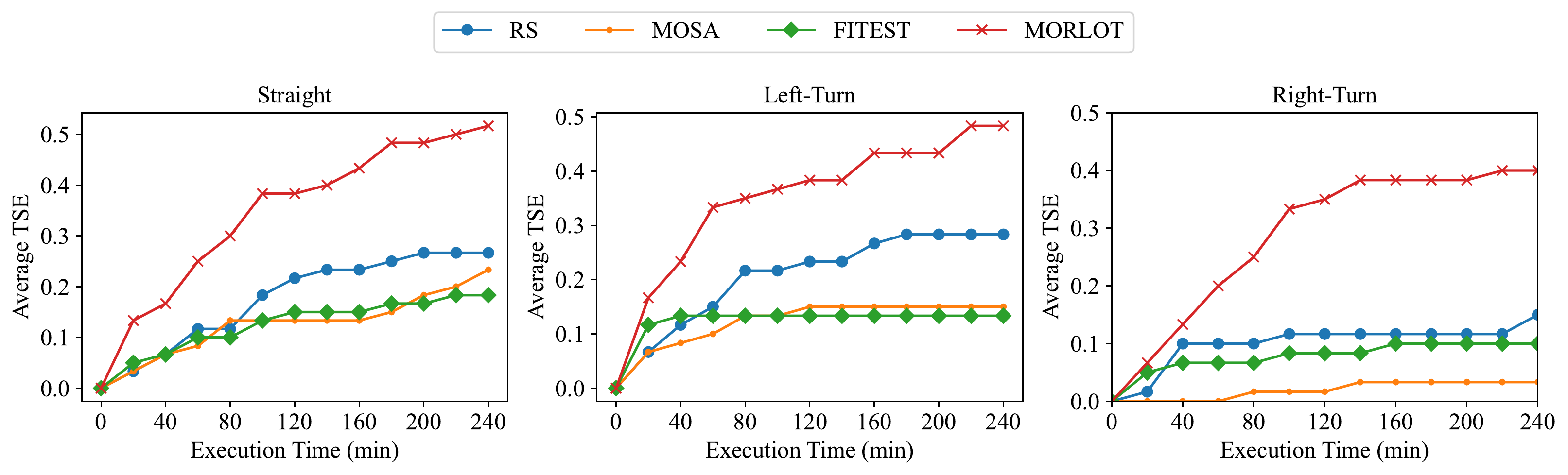}
	\caption{Average \textit{TSE} values over 20 minutes interval}
	\label{fig:efficiency}
\end{figure*}

Figure~\ref{fig:efficiency} shows the relationship between the execution time and the average \textit{TSE} values for 10 runs of RS, MOSA, FITEST, and MORLOT for each of the three initial environments.

Overall, we can clearly see that MORLOT is always at the top in all environments, meaning that MORLOT always achieves the highest \textit{TSE} values, at any time, when compared to the other approaches over the same period of time.
Furthermore, the gaps between MORLOT and the others keep increasing over time, up to a certain point, implying that MORLOT is not only significantly more efficient at detecting unknown violations but also that the effect size becomes larger as we run the testing approaches longer.
An tentative explanation for this observation is that MORLOT keeps learning over time as it observes further states and rewards during the generation of test scenarios. 
Furthermore, as described in Section~\ref{sec:eval:eff:setup}, dynamically decreasing the epsilon value in MORLOT makes it gradually more exploitative and more effective. 

Comparing RS, MOSA, and FITEST, we can see that MOSA and FITEST do not significantly outperform RS at any time, meaning that advanced search-based approaches are not more efficient than simple random search with an archive in this context.
As already discussed in RQ1, this can be mainly because of the enormous search space that makes advanced search approaches less effective within a practical time budget.

\begin{tcolorbox}
The answer to RQ2 is that MORLOT is significantly more efficient than random search and alternative many-objective search approaches. Indeed, it achieves, for any given time budget, a significantly higher average \textit{TSE}, and this difference keeps increasing over time. 
\end{tcolorbox}

\subsection{Threats to Validity}
Using one DADS and one simulator is a potential threat to the external validity of our results.
To mitigate the issue, we selected the highest rank DADS (i.e., Transfuser), at the time of our evaluation, among publicly available ones in the CARLA Automation Driving Leaderboard~\cite{carla-leaderboard}, and a high-fidelity driving simulator (i.e., CARLA) that can be coupled with the selected DADS; they are representative of state-of-the-art DADS and advanced driving simulators, respectively, in terms of performance and fidelity~\cite{CARLA-citation,Prakash2021CVPR}. 
Note that we did not consider more DADS from the Leaderboard since (1) most of them are not good enough to drive the ego vehicle safely (e.g., yielding many violations of the given requirements even by random search) and (2) our evaluation already took more than 600 computing hours. 
Nevertheless, further studies will be needed to increase the generalizability of our results.

The degree of the discretization of the actions and states could be a potential factor that affects our results. 
However, the same set of possible actions (changes) is used for RS, MOSA, FITEST, and MORLOT.
Furthermore, since only MORLOT considers states, using a sub-optimal discretization of these states would only decrease the effectiveness and efficiency of MORLOT, and therefore this aspect has no impact on our conclusions.

Possible actions and states for MORLOT (and other alternative approaches) could affect the realism of the generated test scenarios. In general, we ensure that scenarios are realistic in the sense that they are physically possible. This does not imply they are likely. Indeed, for such DNN-enabled autonomous systems, it is important to test them in a conservative way. For example, if the vehicle in front is coming out of the lane and returning, we consider this behavior to be realistic and representative of an incapacitated driver. To achieve realism, we keep the magnitude of changes small enough to be physically possible, as explained in Section~\ref{sec:eval:eff:setup}.
\section{Conclusion}\label{sec:conclusion}

In this paper, we present MORLOT, a novel approach that combines Reinforcement Learning (RL) and many-objective search to effectively and efficiently generate a test suite for DNN-Enabled Systems (DES) by dynamically changing the application environment. 
We specifically address the issue of scalability when many requirements must be validated. 
We empirically evaluate MORLOT using a state-of-the-art DNN-enabled Automated Driving System (DADS) integrated with a high-fidelity driving simulator. 
The evaluation results show that MORLOT is significantly more effective and efficient, with a large effect size, than random search and many-objective search approaches tailored for test suite generation.

As part of future work, we plan to investigate if we can reuse MORLOT's Q-tables trained on a former version of the DES under test to improve test effectiveness and efficiency for later versions.
We also plan to extend MORLOT to leverage different RL methods, including Deep Q-learning with varying epsilon values, and compare the results in the context of DADS online testing.

\section{Data Availability}\label{sec:data}

The replication package of our experiments, including the implementation of MORLOT and alternative approaches, the instructions to set up and configure the DADS and simulator, the detailed descriptions of the initial environments used in the experiments, and videos of requirement violations found by MORLOT, are publicly available on FigShare~\cite{morlat-supp}.

\bibliographystyle{IEEEtranN}
\bibliography{DNN-Testing-bibliography}

\end{document}